\documentclass[letterpaper]{article} 
\usepackage{aaai25}  
\usepackage{times}  
\usepackage{helvet}  
\usepackage{courier}  
\usepackage[hyphens]{url}  
\usepackage{graphicx} 
\usepackage{natbib}  
\usepackage{caption} 
\usepackage{algorithm}
\usepackage{algorithmic}
\usepackage{amsmath,amsthm,amssymb,amsfonts}
\usepackage{booktabs}
\usepackage{multirow}
\usepackage{multicol}
\usepackage{appendix}

\usepackage{newfloat}
\usepackage{listings}
\DeclareCaptionStyle{ruled}{labelfont=normalfont,labelsep=colon,strut=off} 
\floatstyle{ruled}
\newfloat{listing}{tb}{lst}{}
\floatname{listing}{Listing}
\title{HEP-NAS: Towards Efficient Few-shot Neural Architecture Search via Hierarchical Edge Partitioning}

\author {
    Jianfeng Li\textsuperscript{\rm 1},
    Jiawen Zhang\textsuperscript{\rm 1},
    Feng Wang\textsuperscript{\rm 1}\thanks{Corresponding authors.},
    Lianbo Ma\textsuperscript{\rm 2}\footnotemark[1]
}
\affiliations {
    \textsuperscript{\rm 1}School of Computer Science, Wuhan University\\
    \textsuperscript{\rm 2}College of Software, Northeastern University\\
    fengwang@whu.edu.cn, malb@swc.neu.edu.cn
}

\begin{document}

\maketitle

\begin{abstract}
One-shot methods have significantly advanced the field of neural architecture search (NAS) by adopting weight-sharing strategy to reduce search costs. However, the accuracy of performance estimation can be compromised by co-adaptation. Few-shot methods divide the entire supernet into individual sub-supernets by splitting edge by edge to alleviate this issue, yet neglect relationships among edges and result in performance degradation on huge search space. 
In this paper, we introduce HEP-NAS, a hierarchy-wise partition algorithm designed to further enhance accuracy. To begin with, HEP-NAS treats edges sharing the same end node as a hierarchy, permuting and splitting edges within the same hierarchy to directly search for the optimal operation combination for each intermediate node. This approach aligns more closely with the ultimate goal of NAS. Furthermore, HEP-NAS selects the most promising sub-supernet after each segmentation, progressively narrowing the search space in which the optimal architecture may exist. To improve performance evaluation of sub-supernets, HEP-NAS employs search space mutual distillation, stabilizing the training process and accelerating the convergence of each individual sub-supernet. Within a given budget, HEP-NAS enables the splitting of all edges and gradually searches for architectures with higher accuracy. 
Experimental results across various datasets and search spaces demonstrate the superiority of HEP-NAS compared to state-of-the-art methods. Our code is available at \url{https://github.com/Jianf-l/hepnas}.
\end{abstract}

\section{Introduction}

In recent years, NAS has received widespread attention from both academia and industry \cite{jiang2024meco,wang2023fp,wu2024g} for its ability to automatically search for optimal network architecture for specific tasks. Compared to traditional manually designed network \cite{he2016deep}, NAS simplifies the human iterative design process and uncovers more efficient and innovative network architectures.

NAS was first introduced by \cite{zoph2016neural} using reinforcement learning, which requires substantial computational resources, thereby impeding its practical application. To improve search efficiency, \cite{pham2018efficient} proposed one-shot NAS, which integrates all potential networks within a supernet and trains them simultaneously using a weight-sharing strategy, then uses this supernet as an estimator to evaluate the performance of candidate architectures. This innovation significantly reduces the time cost from thousands of GPU-Days to merely a few.
However, despite notable improvements, one-shot NAS is criticized \cite{bender2018understanding,wang2021rethinking} for its diminished effectiveness as a proxy for evaluating candidate architecture performance and difficulties in identifying superior architectures within the search space, due to the inherent properties of joint training, namely co-adaptation. To elaborate, consider two sub-supernets \(\mathcal{N}_{a}(w_{a},w_{s})\) and \(\mathcal{N}_{b}(w_{b},w_{s})\) that share the same weight \(w_{s}\). If they produce mismatched gradients for \(w_{s}\) when optimized independently, then optimizing the whole supernet \(\mathcal{N}_{s}(w_{a},w_{b};w_{s})\) would lead to a compromise weight aimed at a global optimum, which may not be ideal for either \(\mathcal{N}_{a}\) or \(\mathcal{N}_{b}\) individually. As a result, evaluation of \(\mathcal{N}_{a}\) and \(\mathcal{N}_{b}\) might not reflect their true performance. To improve its accuracy, few-shot methods \cite{zhao2021few,hu2022generalizing,ly2024analyzing} divide operations into distinct sub-supernets via edge-wise partition strategy, facilitating weight-sharing within each sub-supernet while maintaining separation across them, thereby mitigating disagreement among sub-supernets on how to update the weights of shared module, which proved to be effective. 

Although few-shot methods outperform one-shot counterparts in performance, their partition strategy is not ideal. 
Firstly, it overlooks the interrelationships among edges, providing limited alleviation to co-adaption within a hierarchy. To begin with, selecting the optimal combination of operations for each intermediate node, works more relevant to the objectives of NAS than choosing the best operation for each edge independently. However, with edge-wise partition strategy, after segmenting a specific edge, each resulting sub-supernet still retains all operations on other edges within the same hierarchy, leading to co-adaptation within this hierarchy persisting during the subsequent joint training process until all edges in this hierarchy are split. As a result, when assessing the performance of candidate sub-supernets and determining the optimal connections for an intermediate node using weights inherited from the supernet, the performance estimation remains inaccurate since the combination of operations connected to this node is not fully trained in isolation, which can hinder the identification of architectures with truly high accuracy.
Secondly, it maintains unnecessary search spaces, which splits all sub-supernets and constructs a full multifork partition tree, leading to significantly increased time overhead. Therefore, only a limited number of edges are split in most cases to balance search cost, constraining its improvement over one-shot methods. However, not all branches warrant further exploration, as high-performance architectures often favor similar operations on the same edge, rendering many other operations redundant \cite{wan2022redundancy}. Therefore, exploring these search spaces may waste search budgets with little performance benefit.

In this paper, we propose HEP-NAS, a hierarchical edge partitioning algorithm to address above issues. The key concept involves two alternate process. The first one is splitting all edges via hierarchy-wise partition to ensure operation combinations can be trained in isolation. The other one is gradually narrowing the search space to save search cost. Furthermore, as greedily selecting the optimal search space may entail the risk of falling into a local optimum and require sufficient retraining process, we introduce SMD (search space mutual distillation) to ensure the efficacy of search space reduction. Specifically, sub-supernets partitioned within the same hierarchy engage in collaborative learning and mutual teaching to boost the convergence of individual sub-supernets, while the optimal sub-supernet identified in the preceding hierarchy guide all sub-supernets on the current hierarchy to mitigate performance degeneration caused by discretization, thereby offering more accurate selection of the optimal search space.

To evaluate the efficacy of HEP-NAS, we conducted extensive experiments across various datasets and search spaces. The experimental outcomes demonstrate that HEP-NAS outperforms alternative state-of-the-art methods. The discovered architectures not only achieve 97.56\% accuracy on CIFAR-10, but also exhibit a 76.4\% top-1 accuracy when directly transferred to ImageNet, underscoring the exceptional generalization capability of HEP-NAS.

\section{Related Work}

One-shot methods builds one single supernet that includes all candidate architectures in the search space as sub-paths, where architectures can share weights with each other as long as they contain the same operation on the same edge. In the training process, it employs a weight-sharing strategy to optimize the weight of supernet \(\mathcal{W}_{\mathcal{A}}\) only once. Subsequently, in the evaluation process, the sampled architectures treat the weights inherited from the supernet \(w_{\alpha}\) as if they were trained alone and are ranked by their accuracy on the validation set to derive the optimal one. Later, differentiable methods \cite{liu2018darts,chu2020darts,chu2020fair} eschew discrete architecture candidate searches in favor of continuous relaxation of the search space. This enables optimization through efficient gradient descent. 

Despite the search efficiency, one-shot methods often yields suboptimal results due to inaccurate performance estimation of child networks caused by co-adaptation. Various methods have been introduced to improve its accuracy. For example, SPOS \cite{guo2020single} and its successors \cite{chu2023mixpath,chen2023mngnas,zhang2024boosting} construct a supernet with choice blocks, each containing different operations with only one activated during training to mitigate co-adaptation within the blocks. However, this approach gives rise to multi-model forgetting \cite{zhang2020overcoming}, introducing additional complexity. 
Shapley-NAS \cite{xiao2022shapley} uses marginal contribution to assess operation importance, offering insights into their strengths and weaknesses but incurring higher time overhead. 

Few-shot NAS divides the supernet into sub-supernets through edge-wise partitioning, restricting weight-sharing within each sub-supernet while maintain separation across them. 
These sub-supernets are trained in one-shot manner, with the optimal architecture obtained by conducting searching and evaluating process on each of them. 
GM-NAS \cite{hu2022generalizing}, an evolution of few-shot NAS, groups individual operations on each edge using gradient matching, which reduces the number of sub-supernets significantly, yet still splits limited edges due to maintaining unnecessary search spaces. Recent research \cite{ly2024analyzing} extends gradient matching and introduces diverse metrics to group operations within an edge but overlooks correlations among edges. Therefore, it is still required to effectively develop a more precise performance estimator to search for architectures with higher accuracy.

\begin{figure*}[t]
    \centering
    \includegraphics[width=0.9\linewidth]{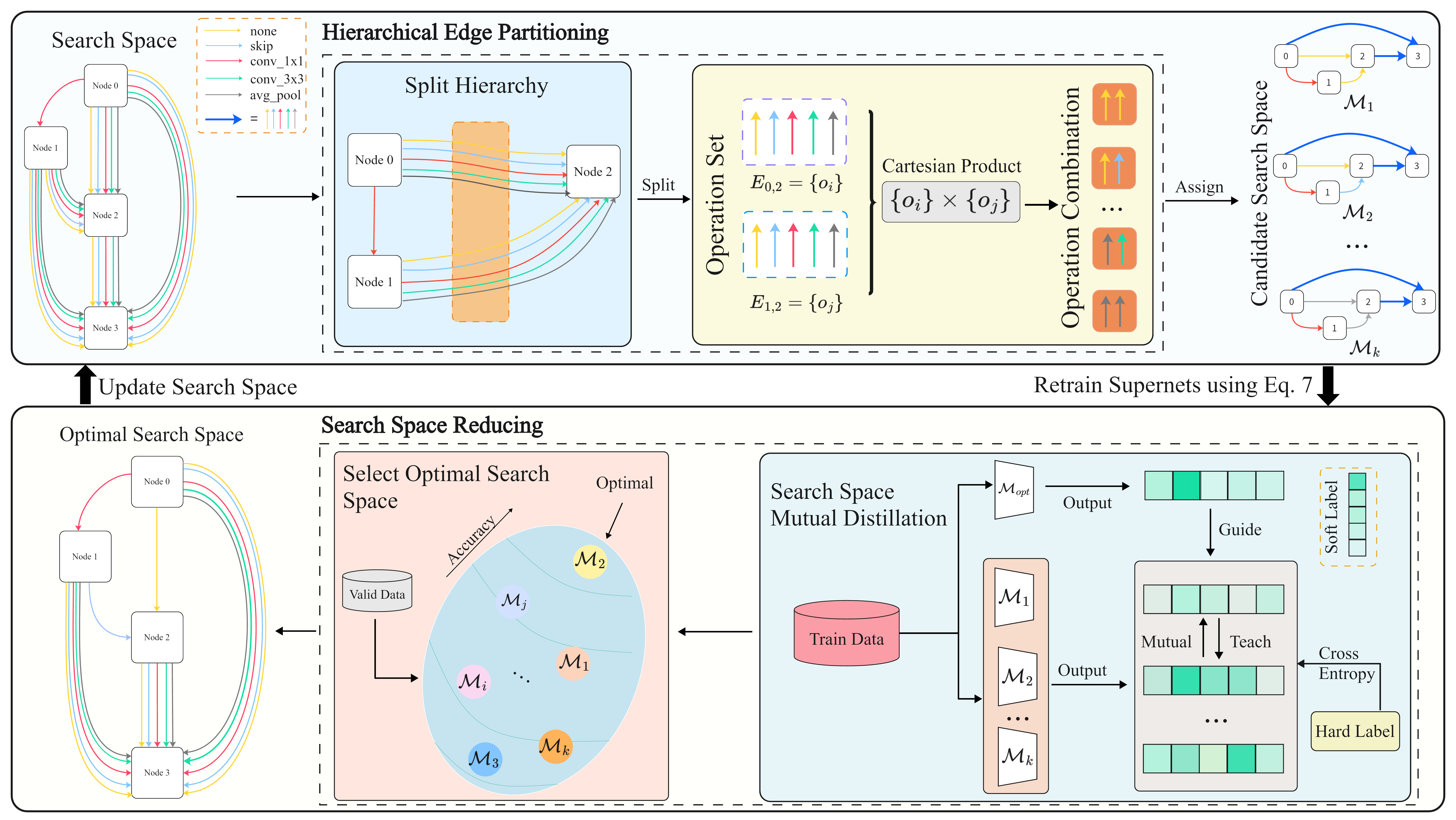}
    \caption{Overall illustration of HEP-NAS. A hierarchy-wise partition strategy is utilized to create sub-supernets. Subsequently, SMD is employed to expedite and stabilize the training process of these sub-supernets. The next step involves selecting the sub-supernet with the highest accuracy on the validation dataset, replacing the previous optimal one, and proceeding to split the next hierarchy until all hierarchies have been partitioned.}
    \label{fig1}
\end{figure*}

\section{Methodology}

\subsection{Preliminaries}

\subsubsection{Gradient Matching} Gradient matching approach proposed by \cite{hu2022generalizing} significantly reduces the number of generated sub-supernets after segmentation through grouping together operations whose standalone gradient is similar. The gradient matching metric introduced to quantify the similarity of operations can be formulated as,
\begin{equation}\label{eq1}
    GM(o_{i}^{k},o_{j}^{k})=\mathcal{S}_{cos}(\nabla\mathcal{L}(\mathcal{M}_{o_{i}}^{k},\omega),\nabla\mathcal{L}(\mathcal{M}_{o_{j}}^{k},\omega))
\end{equation}
where \(\mathcal{S}_{cos}\) represents the cosine similarity function, \(o_{i}^{k}\) is the \textit{i}-th operation on the \textit{k}-th edge \(e_{k}\), and \(\mathcal{M}_{o_{i}}^{k}\) stands for a sub-supernet with only operation \(o_{i}\) left enabled on edge \(e_{k}\). After evaluating the gradient matching score for each pair of operations on \(e_{k}\), groups can be formed via min-cut optimization, defined as,
\begin{equation}\label{eq2}
    \mathcal{G}_{k}=\mathop{\arg \min}_{\mathcal{G}_{k} \subseteq \mathcal{O}} \sum_{o_{i}^{k} \in \mathcal{G}_{k},o_{j}^{k} \in \mathcal{O} \setminus \mathcal{G}_{k}} GM(o_{i}^{k},o_{j}^{k})
\end{equation}
where \(\{\mathcal{G}_{k},\mathcal{O}\setminus\mathcal{G}_{k}\}\) are the obtained operation groups on \(e_{k}\).

We utilize the gradient matching technique to group operations, and progressively search for architectures with higher accuracy via alternate hierarchical edge partitioning and search space reducing process (see Fig. \ref{fig1}). Initially, HEP-NAS trains the entire supernet in a one-shot fashion. It then proceeds, following a hierarchical sequence, to rearrange operations within the current hierarchy, combining and assigning them to distinct sub-supernets. These sub-supernets undergo training using SMD, with the most accurate one on the validation dataset chosen as optimal for further splitting. This iterative process continues for subsequent hierarchies until all hierarchies have been partitioned. Details are present in the following subsections.

\subsection{Hierarchical Edge Partitioning}
Consider an intermediate node \(N_{k}\) within a given search space \(\mathcal{S}\). \(N_{k}\) is connected to its predecessor nodes, where it aggregates, collects, and processes the feature maps passed by them, without involving the connections of the rest of the nodes. This connectivity is expressed as:
\begin{equation}\label{eq3}
    x^{(k)}=\sum_{i<k}\sum_{o \in \mathcal{O}}o^{(i,k)}(x^{(i)})
\end{equation}
Here, \(x^{(k)}\) denotes the output feature map of node \(N_{k}\), and \(o^{(i,k)}\) represents an individual operation on edge \(E_{i,k}\) connecting \(N_{i}\) and \(N_{k}\). Since the ultimate objective is to simultaneously select the optimal operation on each edge \(E_{i,k}\) for \(i=0,1,\cdots,k-1\), we treat these edges connected to the same end node \(N_{k}\) as a hierarchy \(h_{k}\), defined as \(h_{k}=\{E_{i,k}\}_{0 \leq i < k}\), and directly search for the best combination of operations within \(h_{k}\) for \(N_{k}\). To elaborate, let \(\psi^{k}\) represent the operation combinations in \(h_{k}\). \(\psi^{k}\) is calculated by:
\begin{equation}\label{eq4}
    \forall{k}\in\{2,3,\cdots,N-1\},\psi^{k}=\prod_{i<k} \{\mathcal{G}^{(i,k)},\mathcal{O} \setminus \mathcal{G}^{(i,k)}\}
\end{equation}
where \(\prod\) denotes the Cartesian product and \(\mathcal{G}^{(i,k)}\), \(\mathcal{O} \setminus \mathcal{G}^{(i,k)}\) are operation groups on \(E_{i,k}\) calculated by Eq. \ref{eq2}. Instead of splitting one edge and assigning each individual group of operation into separate sub-supernets, we rearrange operation groups within the same hierarchy in combination and allocate these combinations to distinct sub-supernets \(\{\mathcal{M}_{k}\}\) derived from the segmentation of \(h_{k}\), with their edges replaced accordingly and their weights inherited from parent sub-supernet. Therefore, there will be only one operation group reserved on each edge connecting \(N_{k}\) to its predecessor node after segmentation, ensuring each combination can be fully trained in isolation as much as possible to obtain its truly performance.
For instance, the segmentation of \(h_{2}\) (\(\{\{\mathcal{G}^{(0,2)}, \mathcal{O} \setminus \mathcal{G}^{(0,2)}\},\{\mathcal{G}^{(1,2)}, \mathcal{O} \setminus \mathcal{G}^{(1,2)}\}\}\)) will generate \(2^{2}=4\) sub-supernets, with their second hierarchy only containing \(\{\{\mathcal{G}^{(0,2)}\},\{\mathcal{G}^{(1,2)}\}\}\), \(\{\{\mathcal{G}^{(0,2)}\},\{\mathcal{O} \setminus \mathcal{G}^{(1,2)}\}\}\), \(\{\{\mathcal{O} \setminus \mathcal{G}^{(0,2)}\},\{\mathcal{G}^{(1,2)}\}\}\), \(\{\{\mathcal{O} \setminus \mathcal{G}^{(0,2)}\},\{\mathcal{O} \setminus \mathcal{G}^{(1,2)}\}\}\) respectively. 
The hierarchical segmentation strategy establishes a broader and shallower partition tree, with each layer of the tree training the sub-supernets created by the varied connectivity of the corresponding predecessor nodes directly. This approach more effectively addresses co-adaptation at the hierarchical level and searches out architectures with higher accuracy.

Additionally, unlike few-shot methods where edges can be split in any order, hierarchical edge partitioning splits each hierarchy in node numbering sequence. This is because each successive intermediate node is connected to more predecessor nodes than the previous one, and a premature perturbation of a significant number of edges can destabilize the architectural training, resulting in biased outcomes. Therefore, we gradually increase the strength of splitting, ensuring that the segmentation of \(h_{k+1}\) generates more sub-supernets than the segmentation of \(h_{k}\).

\begin{algorithm}[t]
\caption{Main process of HEP-NAS}
\label{alg1}
\textbf{Input}: Search space \(\mathcal{S}\), warmup epochs \(warm\_epo\) to train sub-supernets before selecting the optimal one, the set of each epoch \(split\_epos\) to split the supernet.\\
\textbf{Parameters}: The optimal sub-supernet \(\mathcal{M}_{opt}\) found on previous hierarchy, the set of candidate operations \(\{o^{(i,k)}\}_{i<k}\) on edges connecting the \textit{i}-th node and \textit{k}-th node.\\
\textbf{Output}: The most promising architecture

\begin{algorithmic}[1] 
\STATE \(start\_epo \gets 0\);
\STATE \(k \gets 2\);
\FOR{\(end\_epo\) in \(split\_epos\)}
    \STATE Train \(\mathcal{S}\) for \(end\_epo - start\_epo\) epochs;
    \STATE \(\mathcal{M}_{opt} \gets \mathcal{S}\);
    \STATE Generate combinations \(\psi^{k}\) of \(\{o^{(i,k)}\}_{i<k}\) using Eq. \ref{eq4};
    \STATE  Split \(\mathcal{S}\) on the \textit{k}-th hierarchy and assign each element in \(\psi^{k}\) into sub-supernets \(\{\mathcal{M}_{k}\}\); 
    \FOR{\(i=1,2,\cdots,warm\_epo\)}
        \STATE Sequentially train each sub-supernet \textit{m} in \(\{\mathcal{M}_{k}\}\) for one epoch with the guidance of \(\mathcal{M}_{opt}\) via minimizing Eq. \ref{eq7};
    \ENDFOR
    \STATE Select sub-supernet \(\mathcal{M}^{*}\) with the highest accuracy on validation dataset using Eq. \ref{eq5};
    \STATE \(\mathcal{S} \gets \mathcal{M}^{*}\);
    \STATE \(start\_epo \gets end\_epo\);
    \STATE \(k \gets k+1\);
\ENDFOR
\STATE Derive the final architecture from \(\mathcal{S}\) using corresponding base model selection methods.
\end{algorithmic}
\end{algorithm}

\subsection{Search Space Reducing}
\subsubsection{Selection of Optimal Search Space} Prior to advancing the segmentation process on \(h_{k+1}\), we retrain \(\{\mathcal{M}_{k}\}\) for a few epochs and then identify the optimal sub-supernet \(\mathcal{M}_{opt}\) for continued splitting, while discarding others to save search costs. 
Since the hierarchy-wise splitting strategy generates more sub-supernets in one stage and allows for fuller comparisons, it does not easily fall into a local optimum compared to the edge-wise strategy with pruning. This strategy greatly improves search efficiency by constantly trimming the number of sub-supernets when splitting all edges, while ensuring that the global optimal architecture can be searched as much as possible. In the meantime, by affecting only a few edges in one partitioning stage, gradually increasing the perturbation strength and shrinking search space, HEP-NAS is able to reduce the discretization error \cite{he2024darts} caused by direct one-step selection from the supernet to final architecture, which can significantly change the computation of the feature map and leads to suboptimal results. This means it suffers less from accuracy degradation after discarding unimportant operations in the final model selection stage. 

A straightforward criterion for selection is the accuracy of sub-supernets on the validation dataset,
\begin{equation}\label{eq5}
    \mathcal{M}_{opt}=\mathop{\arg \max} _{\mathcal{M}_{k}} \ Acc_{val}(\mathcal{M}_{k};w_{\mathcal{M}_{k}})
\end{equation}
where \(Acc_{val}\) stands for the top-1 accuracy on validation dataset. However, segregating supernets inevitably disrupts the network structure and alters the computation of feature maps, leading to suboptimal weights for operations in the contracted architecture, thereby impacting sub-supernet performance and evaluation outcomes. Moreover, determining the appropriate number of training epochs for \(\{\mathcal{M}_{k}\}\) poses a challenge, as too few epochs result in inaccurate performance evaluations due to delayed convergence of operations with more parameters, while too many epochs incur significant time overhead. These factors can lead to the selection of a suboptimal search space.

\subsubsection{Search Space Mutual Distillation} To stabilize the training process and boost the convergence of sub-supernets for better performance evaluation, we introduce SMD. Different from previous works commonly introducing a large pretrained model or a third-party model to distill knowledge, which limits flexibility when there are no existing pretrained models, SMD can employ knowledge transfer among segmented sub-supernets in the same hierarchy without requiring an external teacher model, as these sub-supernets can learn collaboratively and teach each other throughout the training process, thus facilitating the convergence and enhancing generalization capabilities of individual architectures. This is realized by appending an objective function \(\mathcal{L}_{dis}\), the cross entropy with the soft target labels, and this cross entropy distills knowledge among candidate sub-supernets, defined as:
\begin{equation}\label{eq6}
    \mathcal{L}_{dis}(\mathcal{M}_{i},\mathcal{M}_{j})=\sum_{c=1}^{C}p_{c}(x;w_{\mathcal{M}_{i}})\log\frac{p_{c}(x;w_{\mathcal{M}_{i}})}{p_{c}(x;w_{\mathcal{M}_{j}})}
\end{equation}
where \(C\) refers to the number of classes, \(p(x;w_{\mathcal{M}_{i}})\) is soft targets generated by a softmax function that converts feature logits to a probability distribution, and \(x\) stands for a batch of given input data.
Meanwhile, to alleviate the performance degradation of sub-supernets caused by splitting operations, SMD also uses the optimal sub-supernet \(\mathcal{M}_{opt}\) searched in the previous hierarchy to guide all the sub-supernets to be trained in the current hierarchy, thereby reducing the difference before and after discretization and stabilizing the training process. To elaborate, let \(\mathcal{M}^{k}_{m}\) be the \textit{m}-th segmented sub-supernet in \(h_{k}\),  \(\mathcal{M}^{k-1}_{opt}\) be the optimal architecture found in the previous hierarchy \(h_{k-1}\), and  \(\mathcal{L}_{cls}\) is the classification loss with the correct labels \(y\), the complete training loss function of  \(\mathcal{M}^{k}_{m}\) can be formulated as:

\begin{equation}\label{eq7}
\begin{split}
\mathcal{L}(\mathcal{M}^{k}_{m})= &\mathcal{L}_{cls}(p(x;w_{\mathcal{M}^{k}_{m}}),y)+{\mathcal{L}}_{dis}(\mathcal{M}^{k}_{m},\mathcal{M}^{k-1}_{opt})\\&+\frac{1}{\mathcal{D}^{k}-1}\sum_{i=0,i\neq m}^{\mathcal{D}^{k}-1}\mathcal{L}_{dis}(\mathcal{M}^{k}_{m},\mathcal{M}^{k}_{i})
\end{split}
\end{equation}
where \(\mathcal{D}^{k}\) stands for the number of sub-supernets in \(h_{k}\) (\(\mathcal{D}^{k}=|\{\mathcal{M}_{k}\}|\)), and the initial weight \(w_{\mathcal{M}_{m}^{k}}\) is inherited from \(w_{\mathcal{M}_{opt}^{k-1}}\) via weight-sharing strategy. 

Algorithm \ref{alg1} summarizes the main process of HEP-NAS. The number of elements in \(split\_epos\) is equal to the number of intermediate nodes since we split all edges with the same end node each time, and the initial value of \textit{k} is the serial number of the first intermediate node. After the segmentation of the last hierarchy, we could derive the most promising architecture via correponding model selection methods (e.g., DARTS) from smaller search space \(\mathcal{S}\).

\section{Experiments and Analysis}

We evaluate the performance of HEP-NAS on the DARTS search space with CIFAR-10, CIFAR-100, and ImageNet for image classifcation, as well as the NAS-Bench-201 search space with CIFAR-10, CIFAR-100, and ImageNet16-120. All experiments are conducted on single NVIDIA RTX 4090 GPU and the results are obtained in 4 independent runs. Experimental results demonstrate that HEP-NAS outperforms the state-of-the-art algorithm in most cases. 

\begin{table}[!ht]
\centering
\small
\setlength{\tabcolsep}{0.9mm}{
\begin{tabular}{cccc}
\toprule
\multirow{2}{*}{Architecture} & Params & Top-1 & Cost \\
&(M) &(\%) &(G\(\cdot\)d) \\
\midrule
ResNet \cite{he2016deep} &1.7 &4.61 &- \\
DenseNet \cite{huang2017densely} &25.6 &3.46 &- \\
\midrule
ProxylessNAS \cite{cai2018proxylessnas} &7.1 &\textbf{2.08} &4.0 \\ 
PC-DARTS \cite{xu2019pc} &3.6 &2.57 &\textbf{0.1} \\
P-DARTS \cite{chen2019progressive} &3.4 &2.50 &0.3 \\
DARTS- \cite{chu2020darts} &3.5 &2.50 &0.4 \\
DARTS-PT \cite{wang2021rethinking} &3.0 &2.61 &0.8 \\
DDPNAS \cite{zheng2023ddpnas} &3.16 &2.59 &0.075 \\
ADARTS \cite{9802692} &2.9 &3.70 &0.2 \\
IS-DARTS \cite{he2024darts} &4.25 &2.56 &0.42 \\
\midrule
MOEA \cite{10059145} &3.0 &2.77 &2.6 \\
MixPath-c \cite{chu2023mixpath} &5.4 &2.6 &0.25 \\
EOFGA \cite{yuan2023effective} &2.11 &2.59 &0.49 \\
EAEPSO \cite{yuan2023particle} &2.94 &2.74 &2.2 \\
\midrule
few-shot DARTS \cite{zhao2021few} &3.6 &2.60 &1.1 \\
GM DARTS \cite{hu2022generalizing} &3.7 &2.46 &1.1 \\
\midrule
HEP-NAS &3.6 &\textbf{2.44} &1.5 \\
\bottomrule
\end{tabular}
}
\caption{Comparison results of HEP-NAS with state-of-the-art methods on CIFAR-10.}
\label{tab1}
\end{table}

\begin{table}[!ht]
\centering
\small
\setlength{\tabcolsep}{1mm}{
\begin{tabular}{cccc}
\toprule
\multirow{2}{*}{Architecture} & Params & Top-1 & Cost\\
&(M) &(\%) &(G\(\cdot\)d) \\
\midrule
ResNet \cite{he2016deep} & 1.7 & 22.10 & - \\
DenseNet \cite{huang2017densely} & 25.6 & 17.18 & -\\
ShuffleNet \cite{zhang2018shufflenet} & 1.06 & 22.86 & - \\
\midrule
GDAS \cite{dong2019searching} & 3.4  & 18.38  & 0.2 \\
P-DARTS \cite{chen2019progressive} & 3.6 & 17.20 & 0.3 \\
PC-DARTS \cite{xu2019pc} & 4.0 & 17.01 & \textbf{0.1} \\
DARTS- \cite{chu2020darts} & 3.3 & 17.51 & 0.4 \\
ADRATS \cite{9802692} &2.9 &17.06 & 0.2 \\
OLES \cite{jiang2024operation} &3.4 &17.30 &0.4\\
\midrule
MOEA \cite{10059145} &5.8 &18.97 &5.2 \\
EOFGA \cite{yuan2023effective} &2.18 &17.23 &0.94\\
EAEPSO \cite{yuan2023particle} &2.94 &16.94 &2.2\\
\midrule
few-shot DARTS \cite{zhao2021few} & 3.4 & 18.59 & 1.3\\
few-shot SNAS \cite{zhao2021few} & 3.3 & 18.39 & 1.3 \\
\midrule
HEP-NAS & 3.6 & \textbf{16.83} &1.6 \\
\bottomrule
\end{tabular} 
}   
\caption{Comparison results of HEP-NAS with state-of-the-art methods on CIFAR-100.}
\label{tab2}
\end{table}

\begin{table}[!ht]
\centering
\small
\setlength{\tabcolsep}{0.8mm}{
\begin{tabular}{cccc}
\toprule
\multirow{2}{*}{Architecture} &Params &Top-1 &Cost\\
&(M) &(\%) &(G\(\cdot\)d)\\
\midrule
ShuffleNetV2 \cite{ma2018shufflenet} & 5      & 25.1 & - \\
MobileNetV3 \cite{howard2019searching} & 7.4    & 23.4 & - \\
\midrule
SNAS \cite{xie2018snas} & 4.3   & 27.3  & 1.5 \\
ProxylessNAS\dag\cite{cai2018proxylessnas} & - & 24.9  & 8.3   \\
GDAS \cite{dong2019searching} & 5.3 & 26.0 & \textbf{0.3} \\
P-DARTS \cite{chen2019progressive} & - & 24.4 & \textbf{0.3} \\
PC-DARTS\dag \ \cite{xu2019pc}  & 5.3 & 24.2 & 3.8 \\
DARTS-\dag \ \cite{chu2020darts} & 4.9 & 23.8 & 4.5 \\
DARTS-PT \cite{wang2021rethinking} & 4.6 & 25.5 & 0.8 \\
OLES \cite{jiang2024operation} &4.7 &24.5 &0.4\\
\midrule
EAEPSO \cite{yuan2023particle} &4.9 &26.9 &4.0\\ 
MOEA \cite{10059145} &4.7 &26.4 &2.6\\
EOFGA\dag \ \cite{yuan2023effective} &5.7 &24.4 &8.0\\
MixPath-A\dag \ \cite{chu2023mixpath} &5.0 &\textbf{23.1} &10.3\\
\midrule
few-shot Proxyless\dag\cite{zhao2021few} & 4.9 &24.1 & 11.7 \\
GM DARTS(2nd) \cite{hu2022generalizing} & 5.1 & 24.5 & 2.7 \\
GM Proxyless\dag \ \cite{hu2022generalizing} &4.9 & 23.4 & 24.9 \\
\midrule
HEP-NAS & 5.1 & \textbf{23.6} & 1.5\\
\bottomrule
\end{tabular} }  \caption{Comparison results of HEP-NAS with state-of-the-art methods on ImageNet. \dag \ means searching directly on ImageNet, otherwise on CIFAR-10.}
\label{tab3}
\end{table}

\begin{table*}[ht]
\centering
\setlength{\tabcolsep}{1mm}{
\begin{tabular}{ccccccc}
\toprule
\multirow{2}{*}{Architecture} & \multicolumn{2}{c}{CIFAR-10} & \multicolumn{2}{c}{CIFAR-100} & \multicolumn{2}{c}{ImageNet16-120} \\
\cmidrule{2-7}
& Valid & Test & Valid & Test & Valid & Test \\
\midrule
REA \cite{real2019regularized} & 91.25$\pm$0.31  & 94.02$\pm$0.31 & 72.28$\pm$0.95 & 72.23$\pm$0.84 & 45.71$\pm$0.77 & 45.77$\pm$0.80  \\
\midrule
ENAS \cite{pham2018efficient} & 39.77$\pm$0.00 & 54.30$\pm$0.00 & 10.23$\pm$0.12 & 10.62$\pm$0.27   & 16.43$\pm$0.00  & 16.32$\pm$0.00  \\
DARTS \cite{liu2018darts}  &  39.77$\pm$0.00 &   54.30$\pm$0.00 & 38.57$\pm$0.00 & 38.97$\pm$0.00 & 18.87$\pm$0.00  & 18.41$\pm$0.00 \\
PC-DARTS \cite{xu2019pc} & 89.96$\pm$0.15 & 93.41$\pm$0.30  & 67.12$\pm$0.39 & 67.48$\pm$0.89  & 40.83$\pm$0.08  & 41.31$\pm$0.22 \\
SPOS \cite{guo2020single} &88.40$\pm$1.07 &92.24$\pm$1.16 &67.8$\pm$2.00 &68.0$\pm$2.25 &39.28$\pm$3.00 &40.28$\pm$3.00 \\
RSPS \cite{li2020random} & 84.16$\pm$1.69 & 87.66$\pm$2.69 & 45.78$\pm$6.33 & 46.60$\pm$6.57 & 31.09$\pm$5.65 & 30.78$\pm$6.12 \\
CyDAS \cite{yu2022cyclic} &\textbf{91.12$\pm$0.44}      & 94.02$\pm$0.31 & 72.12$\pm$1.23  & 71.92$\pm$1.30 & 40.09$\pm$0.61    &  45.51$\pm$0.72 \\
RD-NAS \cite{dong2023rd} &90.44$\pm$0.27 &93.36$\pm$0.04 &70.96$\pm$2.12 &71.83$\pm$1.33 &43.81$\pm$0.09 &44.46$\pm$1.58 \\
DDPNAS \cite{zheng2023ddpnas} &90.12$\pm$0.05 &93.56$\pm$0.02 &70.78$\pm$0.12 &70.91$\pm$0.07 &44.89$\pm$0.29 &46.13$\pm$0.46\\
EG-NAS \cite{cai2024eg} &90.12$\pm$0.05 &93.56$\pm$0.02 &70.78$\pm$0.12 &70.91$\pm$0.07 &44.89$\pm$0.29 &46.13$\pm$0.46\\
OLES \cite{jiang2024operation} &90.88$\pm$0.10 &93.70$\pm$0.15 &70.56$\pm$0.28 &70.40$\pm$0.22 &44.17$\pm$0.49 &43.97$\pm$0.38 \\
\midrule
few-shot DARTS \cite{zhao2021few} & 84.70$\pm$0.44 & 88.55$\pm$0.02 & 70.17$\pm$2.66 & 70.16$\pm$2.87 & 31.16$\pm$3.93 & 30.09$\pm$4.43 \\
GM DARTS \cite{hu2022generalizing}  & 91.03$\pm$0.24 & 93.72$\pm$0.12 & 71.61$\pm$0.62 & 71.83$\pm$0.97 & 42.19$\pm$0.00 & 42.60$\pm$0.00 \\
\midrule
HEP-NAS &91.07$\pm$0.39 & \textbf{93.86$\pm$0.21} & \textbf{73.47$\pm$0.66} & \textbf{73.48$\pm$0.55} & \textbf{46.28$\pm$0.33} &\textbf{46.51$\pm$0.35}\\
\midrule
\textbf{Optimal} &  91.61 & 94.37 & 73.49 & 73.51 & 46.73 & 47.31  \\
\bottomrule
\end{tabular}
}
\caption{Accuracy rates with standard deviation for HEP-NAS on the NAS-Bench-201 search space.}
\label{tab4}
\end{table*}

\subsection{Results on DARTS Search Space}
\subsubsection{Settings} DARTS is a widely used search space for gradient-based NAS algorithm. Candidate operations include zero, skip connection, 3×3\_max pooling, 3×3\_avg pooling, 3×3/5×5\_depth-wise separable convolution, and 3×3/5×5\_dilated separable convolution. All the corresponding hyper-parameter settings are kept the same as DARTS. We train the supernet for 45 epochs with batch size 128 and split the supernet at 15, 25, 35, and 45 epochs respectively. we set \(warmup\_epo\) to 5 initially and decrease it sequentially when splitting the next hierarchy, as the sub-supernets have already been sufficiently trained. We report the params, top-1 error rate and search cost (GPU-Days, referred to G\(\cdot\)d) for each architecture.

\subsubsection{Results} 
\begin{figure}[!ht]
    \centering
    \includegraphics[width=0.9\linewidth]{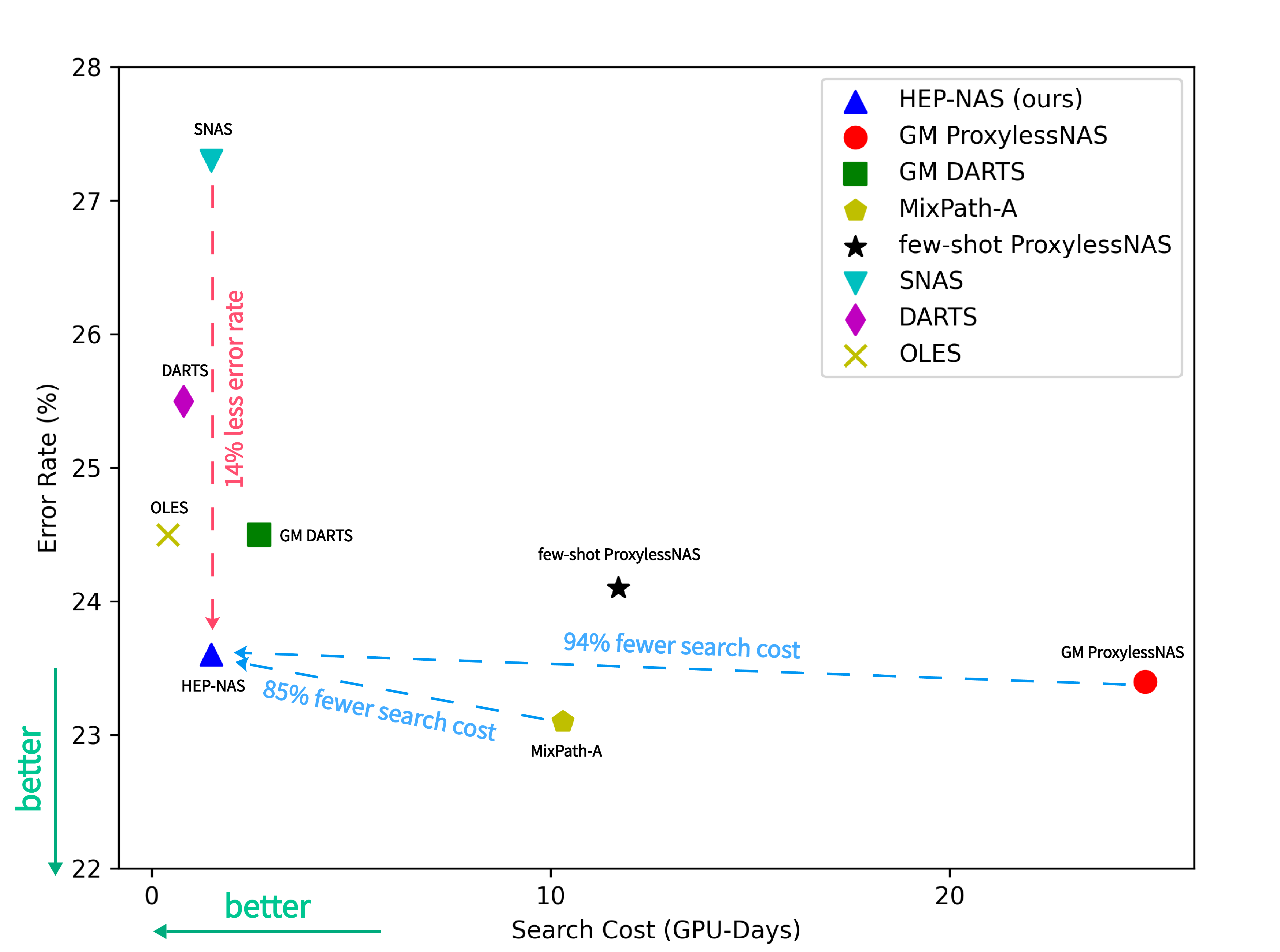}
    \caption{Performance comparation of HEP-NAS with various NAS methods on ImageNet.}
    \label{fig2}
\end{figure}

Results on CIFAR-10 are presented in Table \ref{tab1}. HEP-NAS outperforms few-shot counterparts and other prevailing algorithms with slight additional time overhead. This is achieved by splitting all edges via hierarchy-wise partition strategy to minimize co-adaptation, thereby enhancing evaluation accuracy. Although ProxylessNAS \cite{cai2018proxylessnas} can obtain less error rate, it suffers from a larger parameter size and significantly longer search time. Note that HEP-NAS selects the most promising sub-supernet after each segmentation, therefore additional epochs are required to accurately evaluate performance during search, resulting in a higher cost than original few-shot methods. However, HEP-NAS only keeps one sub-supernet after reduction, allowing direct selection and retraining of the best architecture, while few-shot DARTS and GM DARTS retain all generated sub-supernets and use Successive Halving to progressively discard poor architectures, which requires much more time (almost 40 GPU hours, while HEP-NAS only needs 16) to derive the final well-trained network.
Results on CIFAR-100 are presented in Table \ref{tab2}. HEP-NAS still achieves remarkable performance and surpasses few-shot counterparts in a large margin. Results on ImageNet are presented in Table \ref{tab3}. We directly transfer the architecture searched on CIFAR-10 to the ImageNet dataset to verify its performance. HEP-NAS surpasses most prevailing NAS algorithms, achieving a top-1 test error rate of 23.6\%, which is state-of-the-art when searching architecture directly on CIFAR-10 to save time expenditure, confirming its effectiveness and excellent generalization capability in more complex datasets (see Fig. \ref{fig2}). 

\begin{figure}[!ht]
    \centering
    \includegraphics[width=0.9\linewidth]{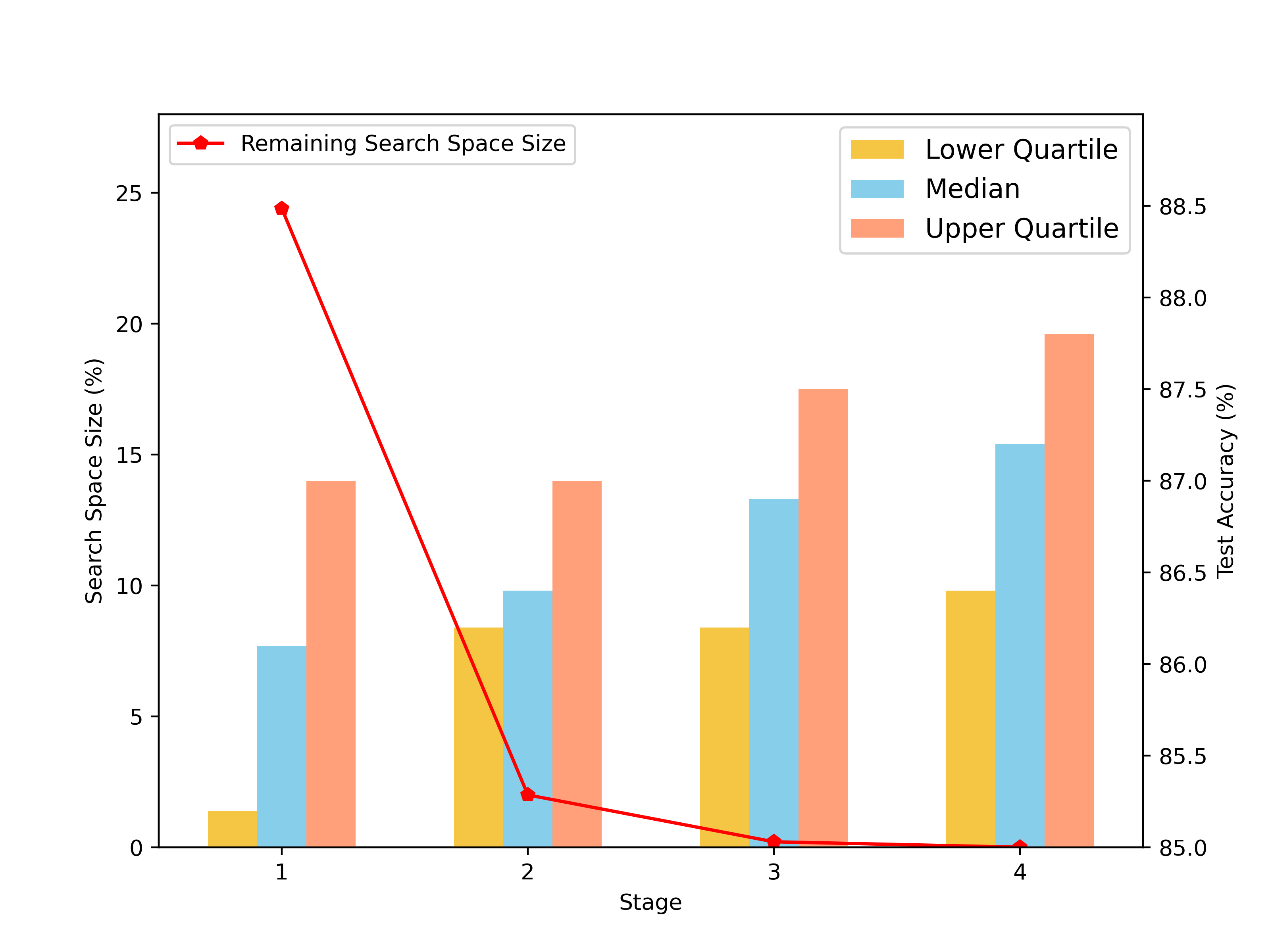}
    \caption{Remaining search space size and accuracy distribution of sub-supernets after each segmentation stage.}
    \label{fig3}
\end{figure}

Fig. \ref{fig3} illustrates the distribution of accuracy rates of sub-supernets and remaining search space size after each segmentation stage. The figure clearly shows a consistent improvement in the accuracy of sub-supernets as the partitioning advances, since we progressively narrow down the search space and employ SMD to enhance the training process, effectively reducing performance degradation. Consequently, this approach provides a more precise evaluation of performance.

\subsection{Results on NAS-Bench-201 Search Space }

\subsubsection{Settings} NAS-Bench-201 \cite{dong2020bench} is another widely used NAS benchmark to analyze the performance of different NAS algorithms, which provides the performance of 15,625 architectures. It is a cell-like search space, including 4 nodes and 5 candidate operations(none, skip, conv\_1x1, conv\_3x3, avgpool\_3x3). We only search architectures on CIFAR-10 and CIFAR-100 , then transfer to ImageNet16-120. The \(warmup\_epo\) is set to 10. We report the mean accuracy rate with standard deviation for each architecture.

\subsubsection{Results}
 The performance of HEP-NAS on NAS-Bench-201 is summarized in Table \ref{tab4}. Compared to few-shot counterparts and other prevailing algorithms, HEP-NAS achieves remarkable performance in most cases by further mitigating co-adaptation to enhance performance evaluation accuracy. Particularly, it achieves an average test accuracy of 46.51\%, ranking highest in the challenging ImageNet16-120 dataset, demonstrating the superiority and generalization capability of HEP-NAS. We also search directly on ImageNet16-120 and obtain 46.34\% accuracy, which is still optimal among counterparts. In addition, HEP-NAS also achieves the highest test accuracy on CIFAR-10 and CIFAR-100, resulting in a 5.31\% improvement over few-shot DARTS on CIFAR-10, and 3.32\% on CIFAR-100 respectively.

 We also evaluate the ranking correlation with Spearman correlation, a particularly important measure to quantify the degree of co-adaption, among architectures in the final reduced search space. We obtain 0.665 Spearman correlation,  proving that HEP-NAS can estimate the performance of architectures accurately.

\subsection{Ablation Study}

\subsubsection{Effectiveness of Hierarchy-wise Partition} Narrowing the search space offers both increased efficiency and risk of converging to a local optimum. By selectively choosing the optimal sub-supernet for continued exploration after splitting each edge, the number of resulting sub-supernets can be reduced further. However, as previously mentioned, the primary goal of the search process is to identify the best combination of operations and determine the most suitable predecessor nodes for each intermediate node. Therefore, considering the connections between nodes is essential. The hierarchy-wise partition strategy accounts for this combinatorial relationship, reducing the likelihood of being trapped in a local optimum and ensuring the exploration of the best overall architecture as far as possible. The experimental results presented in Table \ref{tab5} confirm the efficacy of hierarchical segmentation. 

\begin{table}[ht]
\centering
\setlength{\tabcolsep}{1.5mm}{
\begin{tabular}{cccc}
\toprule
\multirow{2}{*}{Method} & \multicolumn{2}{c}{Error Rate (\%)} & Search Cost\\
\cmidrule{2-3}
& CIFAR-10 & CIFAR-100 & (GPU-Days) \\
\midrule
edge-wise & 2.86 & 17.7 & 0.36 \\
hierarchy-wise & 2.51 & 17.08 & 0.8 \\
\bottomrule
\end{tabular} 
}  
\caption{Ablation study for hierarchy-wise and edge-wise splitting strategy used in conjunction with shrinking search space.}
\label{tab5}
\end{table}

\subsubsection{Order of Hierarchical Segmentation} HEP-NAS employs a hierarchical partitioning strategy, where each successive hierarchy discretizes a larger number of operations than the previous one. This progressive discretization approach is designed to enhance the stability of the training process in the initial phases and accelerate the discovery of the optimal search space in the later stages of training. Therefore, the sequence of hierarchical segmentation is crucial, as premature perturbation of a significant number of edges can destabilize the architectural training, resulting in biased outcomes. To assess the impact of the splitting order, we conducted experiments using reverse order and random order with a batch size of 256. The results are detailed in Table \ref{tab6}.

\begin{table}[!ht]
    \centering
    \setlength{\tabcolsep}{1.5mm}{
    \begin{tabular}{cccc}
    \toprule
    \multirow{2}{*}{Split Order} & \multicolumn{2}{c}{Error Rate (\%)} & Search Cost\\
    \cmidrule{2-3}
    & CIFAR-10 & CIFAR-100 & (GPU-Days) \\
    \midrule
    random order &2.63 &17.08 &1.125 \\
    reverse order &2.68 &17.17 &1.4 \\
    \bottomrule
    \end{tabular}}
    \caption{Ablation Study for splitting order. Results of random order are obtained in 6 independent runs.}
    \label{tab6}
\end{table}

\subsubsection{Effectiveness of Search Space Mutual Distillation} To further confirm the impact of SMD on the training process of sub-supernets, we evaluated four distillation methods: (A) Training sub-supernets without guidance (baseline); (B) Training current sub-supernets using only the previous optimal architecture's guidance; (C) Training current sub-supernets using only guidance from other sub-supernets in the same hierarchy; (D) Training current sub-supernets using guidance from both the previous optimal architecture and other sub-supernets in the same hierarchy, which represents our HEP-NAS algorithm. Experiment results are summarized in Table \ref{tab7}. Method-D boosts the convergence and stabilizes the training process, thereby achieved the lowest test error rate among these methods, showcasing the efficacy of SMD.

\begin{table}[ht]
\centering
\setlength{\tabcolsep}{1mm}{
\begin{tabular}{cccc}
\toprule
\multicolumn{2}{c}{Guidance}   & \multicolumn{2}{c}{Error Rate (\%)} \\
\midrule
previous & other sub-supernets & CIFAR-10& CIFAR-100\\
\midrule
- &- &2.51 & 17.10   \\
\checkmark &- &2.46 &17.10   \\
- &\checkmark  &2.44 &17.07     \\
\checkmark &\checkmark  &2.44 &16.83   \\
\bottomrule
\end{tabular}}
\caption{Ablation study for different distillation method. 'previous' refers to the optimal architecture found in previous hierarchy, and 'other sub-supernets' refers to candidate sub-supernets in current splitting hierarchy.}
\label{tab7}
\end{table}

\subsubsection{Number of hierarchies segmented}
As pointed out in few-shot methods, splitting more edges can ease co-adaption more efficiently, leading to greater accuracy gains, but it can also take longer. We also evaluated the search cost and error rate when splitting partial hierarchies, on CIFAR-100 dataset using DARTS search space, and summarize results in Table \ref{tab8}. From the table, we can see that splitting part of hierarchies can still achieve competitive results with much less time overhead.

\begin{table}[ht]
\centering
\setlength{\tabcolsep}{1mm}{
\begin{tabular}{cccc}
\toprule
Number & Error Rate (\%) & Search Cost (GPU-Days) \\
\midrule
1 &17.2 &0.38 \\
2 &16.88 &0.46 \\
3 &16.85 &0.7 \\
\bottomrule
\end{tabular}}
\caption{Test error and search cost when splitting different number of hierarchies.}
\label{tab8}
\end{table}

\section{Conclusion}
In this paper, we introduce HEP-NAS to search out architectures with higher accuracy. The core concept involves utilizing a hierarchy-wise strategy to partition all edges and progressively narrowing the search space to identify the optimal sub-supernet. To enhance performance evaluation, we concurrently transfer knowledge from the optimal sub-supernet identified in the previous hierarchy and other candidate sub-supernets in the current hierarchy to stabilize the training process and improve convergence of each individual sub-supernet. Extensive experiments across diverse datasets and search spaces demonstrate that HEP-NAS outperforms previous state-of-the-art NAS algorithms in most scenarios.

\section{Acknowledgments} 
This work is supported by the National Nature Science Foundation of China [Grant Nos. 62173258, 61773296].

\bibliography{aaai25}

\begin{thebibliography}{45}
\providecommand{\natexlab}[1]{#1}

\bibitem[{Bender et~al.(2018)Bender, Kindermans, Zoph, Vasudevan, and Le}]{bender2018understanding}
Bender, G.; Kindermans, P.-J.; Zoph, B.; Vasudevan, V.; and Le, Q. 2018.
\newblock Understanding and simplifying one-shot architecture search.
\newblock In \emph{International conference on machine learning}, 550--559. PMLR.

\bibitem[{Cai, Zhu, and Han(2018)}]{cai2018proxylessnas}
Cai, H.; Zhu, L.; and Han, S. 2018.
\newblock Proxylessnas: Direct neural architecture search on target task and hardware.
\newblock \emph{arXiv preprint arXiv:1812.00332}.

\bibitem[{Cai et~al.(2024)Cai, Chen, Liu, Ling, and Lai}]{cai2024eg}
Cai, Z.; Chen, L.; Liu, P.; Ling, T.; and Lai, Y. 2024.
\newblock EG-NAS: Neural Architecture Search with Fast Evolutionary Exploration.
\newblock In \emph{Proceedings of the AAAI Conference on Artificial Intelligence}, volume~38, 11159--11167.

\bibitem[{Chen et~al.(2019)Chen, Xie, Wu, and Tian}]{chen2019progressive}
Chen, X.; Xie, L.; Wu, J.; and Tian, Q. 2019.
\newblock Progressive differentiable architecture search: Bridging the depth gap between search and evaluation.
\newblock In \emph{Proceedings of the IEEE/CVF international conference on computer vision}, 1294--1303.

\bibitem[{Chen et~al.(2023)Chen, Qiu, Li, Zhu, Yang, and Sheng}]{chen2023mngnas}
Chen, Z.; Qiu, G.; Li, P.; Zhu, L.; Yang, X.; and Sheng, B. 2023.
\newblock Mngnas: distilling adaptive combination of multiple searched networks for one-shot neural architecture search.
\newblock \emph{IEEE Transactions on Pattern Analysis and Machine Intelligence}.

\bibitem[{Chu et~al.(2023)Chu, Lu, Li, and Zhang}]{chu2023mixpath}
Chu, X.; Lu, S.; Li, X.; and Zhang, B. 2023.
\newblock Mixpath: A unified approach for one-shot neural architecture search.
\newblock In \emph{Proceedings of the IEEE/CVF International Conference on Computer Vision}, 5972--5981.

\bibitem[{Chu et~al.(2020{\natexlab{a}})Chu, Wang, Zhang, Lu, Wei, and Yan}]{chu2020darts}
Chu, X.; Wang, X.; Zhang, B.; Lu, S.; Wei, X.; and Yan, J. 2020{\natexlab{a}}.
\newblock Darts-: robustly stepping out of performance collapse without indicators.
\newblock \emph{arXiv preprint arXiv:2009.01027}.

\bibitem[{Chu et~al.(2020{\natexlab{b}})Chu, Zhou, Zhang, and Li}]{chu2020fair}
Chu, X.; Zhou, T.; Zhang, B.; and Li, J. 2020{\natexlab{b}}.
\newblock Fair darts: Eliminating unfair advantages in differentiable architecture search.
\newblock In \emph{European conference on computer vision}, 465--480. Springer.

\bibitem[{Dong et~al.(2023)Dong, Niu, Li, Tian, Wang, Wei, Pan, and Li}]{dong2023rd}
Dong, P.; Niu, X.; Li, L.; Tian, Z.; Wang, X.; Wei, Z.; Pan, H.; and Li, D. 2023.
\newblock RD-NAS: Enhancing one-shot supernet ranking ability via ranking distillation from zero-cost proxies.
\newblock In \emph{ICASSP 2023-2023 IEEE International Conference on Acoustics, Speech and Signal Processing (ICASSP)}, 1--5. IEEE.

\bibitem[{Dong and Yang(2019)}]{dong2019searching}
Dong, X.; and Yang, Y. 2019.
\newblock Searching for a robust neural architecture in four gpu hours.
\newblock In \emph{Proceedings of the IEEE/CVF conference on computer vision and pattern recognition}, 1761--1770.

\bibitem[{Dong and Yang(2020)}]{dong2020bench}
Dong, X.; and Yang, Y. 2020.
\newblock Nas-bench-201: Extending the scope of reproducible neural architecture search.
\newblock \emph{arXiv preprint arXiv:2001.00326}.

\bibitem[{Guo et~al.(2020)Guo, Zhang, Mu, Heng, Liu, Wei, and Sun}]{guo2020single}
Guo, Z.; Zhang, X.; Mu, H.; Heng, W.; Liu, Z.; Wei, Y.; and Sun, J. 2020.
\newblock Single path one-shot neural architecture search with uniform sampling.
\newblock In \emph{Computer Vision--ECCV 2020: 16th European Conference, Glasgow, UK, August 23--28, 2020, Proceedings, Part XVI 16}, 544--560. Springer.

\bibitem[{He et~al.(2024)He, Liu, Zhang, and Zheng}]{he2024darts}
He, H.; Liu, L.; Zhang, H.; and Zheng, N. 2024.
\newblock IS-DARTS: Stabilizing DARTS through Precise Measurement on Candidate Importance.
\newblock In \emph{Proceedings of the AAAI Conference on Artificial Intelligence}, volume~38, 12367--12375.

\bibitem[{He et~al.(2016)He, Zhang, Ren, and Sun}]{he2016deep}
He, K.; Zhang, X.; Ren, S.; and Sun, J. 2016.
\newblock Deep residual learning for image recognition.
\newblock In \emph{Proceedings of the IEEE conference on computer vision and pattern recognition}, 770--778.

\bibitem[{Howard et~al.(2019)Howard, Sandler, Chu, Chen, Chen, Tan, Wang, Zhu, Pang, Vasudevan et~al.}]{howard2019searching}
Howard, A.; Sandler, M.; Chu, G.; Chen, L.-C.; Chen, B.; Tan, M.; Wang, W.; Zhu, Y.; Pang, R.; Vasudevan, V.; et~al. 2019.
\newblock Searching for mobilenetv3.
\newblock In \emph{Proceedings of the IEEE/CVF international conference on computer vision}, 1314--1324.

\bibitem[{Hu et~al.(2022)Hu, Wang, Hong, Li, Hsieh, and Feng}]{hu2022generalizing}
Hu, S.; Wang, R.; Hong, L.; Li, Z.; Hsieh, C.-J.; and Feng, J. 2022.
\newblock Generalizing few-shot nas with gradient matching.
\newblock \emph{arXiv preprint arXiv:2203.15207}.

\bibitem[{Huang et~al.(2017)Huang, Liu, Van Der~Maaten, and Weinberger}]{huang2017densely}
Huang, G.; Liu, Z.; Van Der~Maaten, L.; and Weinberger, K.~Q. 2017.
\newblock Densely connected convolutional networks.
\newblock In \emph{Proceedings of the IEEE conference on computer vision and pattern recognition}, 4700--4708.

\bibitem[{Jiang et~al.(2024)Jiang, Ji, Zhu, Yuan, and Huang}]{jiang2024operation}
Jiang, S.; Ji, Z.; Zhu, G.; Yuan, C.; and Huang, Y. 2024.
\newblock Operation-level early stopping for robustifying differentiable NAS.
\newblock \emph{Advances in Neural Information Processing Systems}, 36.

\bibitem[{Jiang, Wang, and Bie(2024)}]{jiang2024meco}
Jiang, T.; Wang, H.; and Bie, R. 2024.
\newblock MeCo: zero-shot NAS with one data and single forward pass via minimum eigenvalue of correlation.
\newblock \emph{Advances in Neural Information Processing Systems}, 36.

\bibitem[{Li et~al.(2023)Li, Yang, Bhardwaj, and Marculescu}]{li2023zico}
Li, G.; Yang, Y.; Bhardwaj, K.; and Marculescu, R. 2023.
\newblock Zico: Zero-shot nas via inverse coefficient of variation on gradients.
\newblock \emph{arXiv preprint arXiv:2301.11300}.

\bibitem[{Li and Talwalkar(2020)}]{li2020random}
Li, L.; and Talwalkar, A. 2020.
\newblock Random search and reproducibility for neural architecture search.
\newblock In \emph{Uncertainty in artificial intelligence}, 367--377. PMLR.

\bibitem[{Lin et~al.(2021)Lin, Wang, Sun, Chen, Sun, Qian, Li, and Jin}]{lin2021zen}
Lin, M.; Wang, P.; Sun, Z.; Chen, H.; Sun, X.; Qian, Q.; Li, H.; and Jin, R. 2021.
\newblock Zen-nas: A zero-shot nas for high-performance image recognition.
\newblock In \emph{Proceedings of the IEEE/CVF International Conference on Computer Vision}, 347--356.

\bibitem[{Liu, Simonyan, and Yang(2018)}]{liu2018darts}
Liu, H.; Simonyan, K.; and Yang, Y. 2018.
\newblock Darts: Differentiable architecture search.
\newblock \emph{arXiv preprint arXiv:1806.09055}.

\bibitem[{Ly-Manson et~al.(2024)Ly-Manson, Leonardon, El~Bey, Hacene, and Mauch}]{ly2024analyzing}
Ly-Manson, T.; Leonardon, M.; El~Bey, A.~A.; Hacene, G.~B.; and Mauch, L. 2024.
\newblock Analyzing Few-Shot Neural Architecture Search in a Metric-Driven Framework.

\bibitem[{Ma et~al.(2018)Ma, Zhang, Zheng, and Sun}]{ma2018shufflenet}
Ma, N.; Zhang, X.; Zheng, H.-T.; and Sun, J. 2018.
\newblock Shufflenet v2: Practical guidelines for efficient cnn architecture design.
\newblock In \emph{Proceedings of the European conference on computer vision (ECCV)}, 116--131.

\bibitem[{Pham et~al.(2018)Pham, Guan, Zoph, Le, and Dean}]{pham2018efficient}
Pham, H.; Guan, M.; Zoph, B.; Le, Q.; and Dean, J. 2018.
\newblock Efficient neural architecture search via parameters sharing.
\newblock In \emph{International conference on machine learning}, 4095--4104. PMLR.

\bibitem[{Real et~al.(2019)Real, Aggarwal, Huang, and Le}]{real2019regularized}
Real, E.; Aggarwal, A.; Huang, Y.; and Le, Q.~V. 2019.
\newblock Regularized evolution for image classifier architecture search.
\newblock In \emph{Proceedings of the aaai conference on artificial intelligence}, volume~33, 4780--4789.

\bibitem[{Wan et~al.(2022)Wan, Ru, Esperan{\c{c}}a, and Li}]{wan2022redundancy}
Wan, X.; Ru, B.; Esperan{\c{c}}a, P.~M.; and Li, Z. 2022.
\newblock On redundancy and diversity in cell-based neural architecture search.
\newblock \emph{arXiv preprint arXiv:2203.08887}.

\bibitem[{Wang et~al.(2021)Wang, Cheng, Chen, Tang, and Hsieh}]{wang2021rethinking}
Wang, R.; Cheng, M.; Chen, X.; Tang, X.; and Hsieh, C.-J. 2021.
\newblock Rethinking architecture selection in differentiable NAS.
\newblock \emph{arXiv preprint arXiv:2108.04392}.

\bibitem[{Wang et~al.(2023)Wang, Zhang, Cui, Yin, and Zhang}]{wang2023fp}
Wang, W.; Zhang, X.; Cui, H.; Yin, H.; and Zhang, Y. 2023.
\newblock FP-DARTS: Fast parallel differentiable neural architecture search for image classification.
\newblock \emph{Pattern Recognition}, 136: 109193.

\bibitem[{Wu et~al.(2024)Wu, Gao, Hong, Wang, Zhou, and Ye}]{wu2024g}
Wu, F.; Gao, J.; Hong, L.; Wang, X.; Zhou, C.; and Ye, N. 2024.
\newblock G-NAS: Generalizable Neural Architecture Search for Single Domain Generalization Object Detection.
\newblock In \emph{Proceedings of the AAAI Conference on Artificial Intelligence}, volume~38, 5958--5966.

\bibitem[{Xiao et~al.(2022)Xiao, Wang, Zhu, Zhou, and Lu}]{xiao2022shapley}
Xiao, H.; Wang, Z.; Zhu, Z.; Zhou, J.; and Lu, J. 2022.
\newblock Shapley-NAS: discovering operation contribution for neural architecture search.
\newblock In \emph{Proceedings of the IEEE/CVF conference on computer vision and pattern recognition}, 11892--11901.

\bibitem[{Xie et~al.(2018)Xie, Zheng, Liu, and Lin}]{xie2018snas}
Xie, S.; Zheng, H.; Liu, C.; and Lin, L. 2018.
\newblock SNAS: stochastic neural architecture search.
\newblock \emph{arXiv preprint arXiv:1812.09926}.

\bibitem[{Xu et~al.(2019)Xu, Xie, Zhang, Chen, Qi, Tian, and Xiong}]{xu2019pc}
Xu, Y.; Xie, L.; Zhang, X.; Chen, X.; Qi, G.-J.; Tian, Q.; and Xiong, H. 2019.
\newblock Pc-darts: Partial channel connections for memory-efficient architecture search.
\newblock \emph{arXiv preprint arXiv:1907.05737}.

\bibitem[{Xue, Chen, and Słowik(2023)}]{10059145}
Xue, Y.; Chen, C.; and Słowik, A. 2023.
\newblock Neural Architecture Search Based on a Multi-Objective Evolutionary Algorithm With Probability Stack.
\newblock \emph{IEEE Transactions on Evolutionary Computation}, 27(4): 778--786.

\bibitem[{Xue and Qin(2023)}]{9802692}
Xue, Y.; and Qin, J. 2023.
\newblock Partial Connection Based on Channel Attention for Differentiable Neural Architecture Search.
\newblock \emph{IEEE Transactions on Industrial Informatics}, 19(5): 6804--6813.

\bibitem[{Yu et~al.(2022)Yu, Peng, Huang, Fu, Du, Wang, and Ling}]{yu2022cyclic}
Yu, H.; Peng, H.; Huang, Y.; Fu, J.; Du, H.; Wang, L.; and Ling, H. 2022.
\newblock Cyclic differentiable architecture search.
\newblock \emph{IEEE transactions on pattern analysis and machine intelligence}, 45(1): 211--228.

\bibitem[{Yuan et~al.(2023)Yuan, Wang, Xue, and Zhang}]{yuan2023particle}
Yuan, G.; Wang, B.; Xue, B.; and Zhang, M. 2023.
\newblock Particle swarm optimization for efficiently evolving deep convolutional neural networks using an autoencoder-based encoding strategy.
\newblock \emph{IEEE Transactions on Evolutionary Computation}.

\bibitem[{Yuan, Xue, and Zhang(2023)}]{yuan2023effective}
Yuan, G.; Xue, B.; and Zhang, M. 2023.
\newblock An effective one-shot neural architecture search method with supernet fine-tuning for image classification.
\newblock In \emph{Proceedings of the Genetic and Evolutionary Computation Conference}, 615--623.

\bibitem[{Zhang et~al.(2024)Zhang, Wang, Qin, and Yan}]{zhang2024boosting}
Zhang, B.; Wang, X.; Qin, X.; and Yan, J. 2024.
\newblock Boosting Order-Preserving and Transferability for Neural Architecture Search: a Joint Architecture Refined Search and Fine-tuning Approach.
\newblock In \emph{Proceedings of the IEEE/CVF Conference on Computer Vision and Pattern Recognition}, 5662--5671.

\bibitem[{Zhang et~al.(2020)Zhang, Li, Pan, Chang, and Su}]{zhang2020overcoming}
Zhang, M.; Li, H.; Pan, S.; Chang, X.; and Su, S. 2020.
\newblock Overcoming multi-model forgetting in one-shot NAS with diversity maximization.
\newblock In \emph{Proceedings of the ieee/cvf conference on computer vision and pattern recognition}, 7809--7818.

\bibitem[{Zhang et~al.(2018)Zhang, Zhou, Lin, and Sun}]{zhang2018shufflenet}
Zhang, X.; Zhou, X.; Lin, M.; and Sun, J. 2018.
\newblock Shufflenet: An extremely efficient convolutional neural network for mobile devices.
\newblock In \emph{Proceedings of the IEEE conference on computer vision and pattern recognition}, 6848--6856.

\bibitem[{Zhao et~al.(2021)Zhao, Wang, Tian, Fonseca, and Guo}]{zhao2021few}
Zhao, Y.; Wang, L.; Tian, Y.; Fonseca, R.; and Guo, T. 2021.
\newblock Few-shot neural architecture search.
\newblock In \emph{International Conference on Machine Learning}, 12707--12718. PMLR.

\bibitem[{Zheng et~al.(2023)Zheng, Yang, Zhang, Wang, Zhang, Wu, Wu, Shao, and Ji}]{zheng2023ddpnas}
Zheng, X.; Yang, C.; Zhang, S.; Wang, Y.; Zhang, B.; Wu, Y.; Wu, Y.; Shao, L.; and Ji, R. 2023.
\newblock Ddpnas: Efficient neural architecture search via dynamic distribution pruning.
\newblock \emph{International Journal of Computer Vision}, 131(5): 1234--1249.

\bibitem[{Zoph and Le(2016)}]{zoph2016neural}
Zoph, B.; and Le, Q.~V. 2016.
\newblock Neural architecture search with reinforcement learning.
\newblock \emph{arXiv preprint arXiv:1611.01578}.

\end{thebibliography}

\clearpage

\section{Appendix}

\subsection{Implementation Details}

\subsubsection{DARTS Search Space} The supernet in DARTS search space consists of 8 cells, where the third and sixth cells are the reduction cells used to generate feature maps with smaller size but more channels. Each cell has 2 inputs (the outputs of the previous cell and the cell before the previous cell) and 4 intermediate nodes connected with 14 edges. 

In the supernet training process, the training dataset is divided into two halves, half for optimize the network weights \(w\) and half for optimize architecture parameters \(\alpha\). We use momentum SGD to optimize the network weights and clip the gradient by 5, with initial learning rate \(\eta_{w}=0.025\) (annealed down to 0.001 following a cosine schedule without restart), momentum 0.9, and weight decay \(3\times10^{-4}\). We use Adam as the optimizer for architecture parameters, with initial learning rate \(\eta_{\alpha}=3\times10^{-4}\), momentum \(\beta=(0.5, 0.999)\) and weight decay \(10^{-3}\). The supernet is trained with these settings for 45 epochs with a fixed \(\alpha\) for the first 5 epochs, and is split into individual sub-supernets via hierarchical edge partitioning at 15, 25, 35, 45 epochs respectively. After each segmentation, we train these sub-supernets with SMD for decreasing epochs (5 epochs for the first hierarchy, then 4 epochs for the second hierarchy) and select the sub-supernet with the highest accuracy on validation dataset for further segmentation. Therefore, after splitting all hierarchies, there only exists one supernet, and we directly select the optimal architecture based on learned \(\alpha\) like DARTS as the final architecture. The search process is conducted on CIFAR-10 and CIFAR-100 separately in four independent runs, while on ImageNet, we directly transfer the searched architecture on CIFAR-10 for evaluation.

In the architecture evaluation process, we follow the retrain settings of DARTS for fair competition. On CIFAR-10 and CIFAR-100, we stack 20 cells to compose the final derived architecture and set the initial channel number as 36. The derived architecture is trained from scratch with a batch size 96 for 600 epochs. We use SGD with an initial learning rate of 0.025, a momentum of 0.9, a weight decay of \(3\times10^{-4}\), and a cosine learning rate scheduler. In addition, we also employ the cutout regularization with length 16, drop-path with probability 0.2, and an auxiliary tower of weight 0.4. On ImageNet, we stack 14 cells (the same as cells searched on CIFAR-10) to compose the final architecture and set the initial channel number as 48. Data augmentation includes random cropping, color jittering, random horizontal flipping and normalization. The architecture is trained for 250 epochs with a batch size of 128, a momentum of 0.9 and a weight decay of \(3\times10^{-5}\). The learning rate decay from 0.1 to 0 following a cosine scheduler.

\subsubsection{NAS-Bench-201 Search Space} 
The supernet in NAS-Bench-201 is stacked by 5 cells and each cell is constructed by 4 nodes connected with 6 edges. We train the supernet for 30 epochs with a fixed \(\alpha\) for the first 10 epochs and split the supernet at 10, 20, 30 epochs respectively. After each segmentation, we train the generated sub-supernets with SMD for 10 epochs, and select the sub-supernet with the highest accuracy on validation dataset for further splitting. Other supernet training settings are kept the same as settings in DARTS search space. After searching out the best architecture, we call the NAS-Bench-201 API to directly obtain accuracy information.

\subsection{Searched Cells}
\begin{figure}[!ht]
    \centering
    \includegraphics[width=0.8\linewidth]{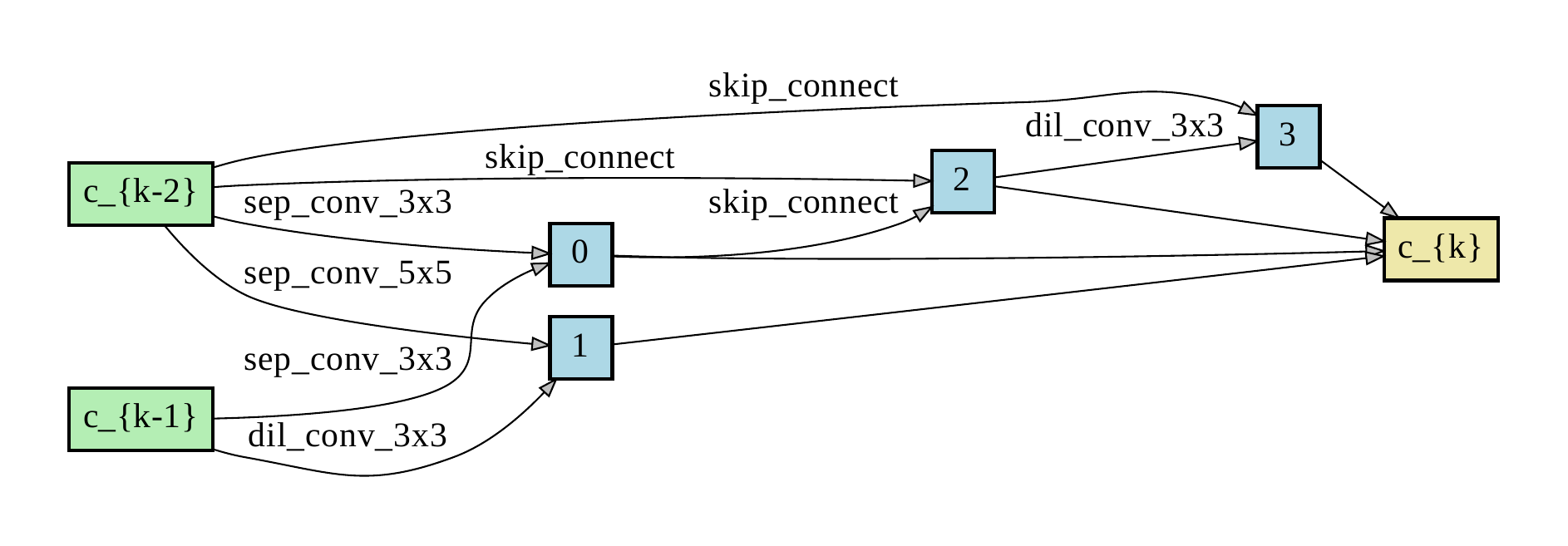}
    \caption{Normal cell on CIFAR-10.}
    \label{fig: cifar10_noraml}
\end{figure}

\begin{figure}[!ht]
    \centering
    \includegraphics[width=0.8\linewidth]{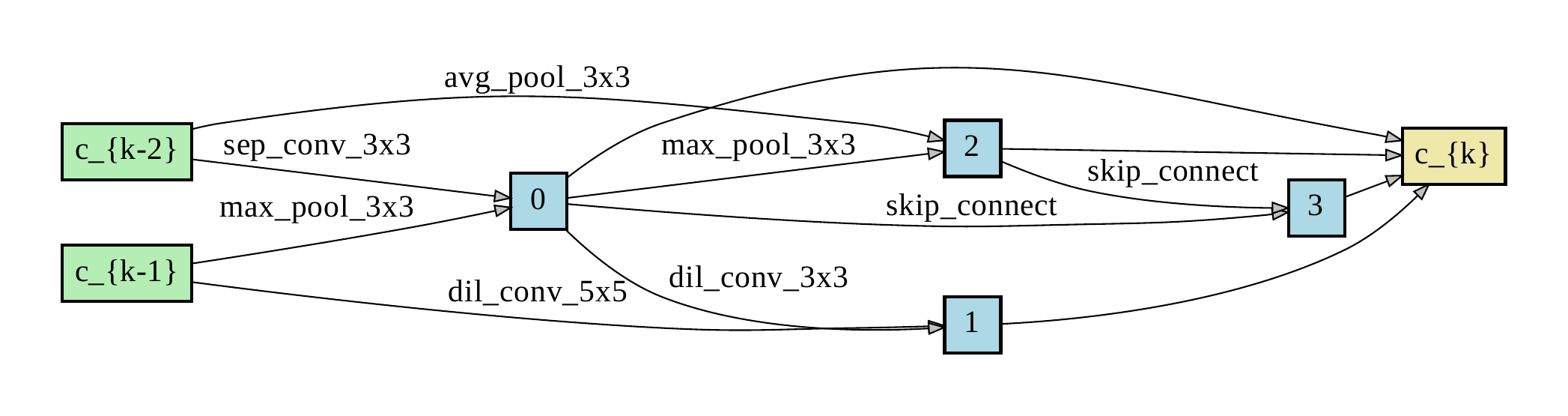}
    \caption{Reduction cell on CIFAR-10.}
    \label{fig: cifar10_reduction}
\end{figure}

\begin{figure}[!ht]
    \centering
    \includegraphics[width=0.8\linewidth]{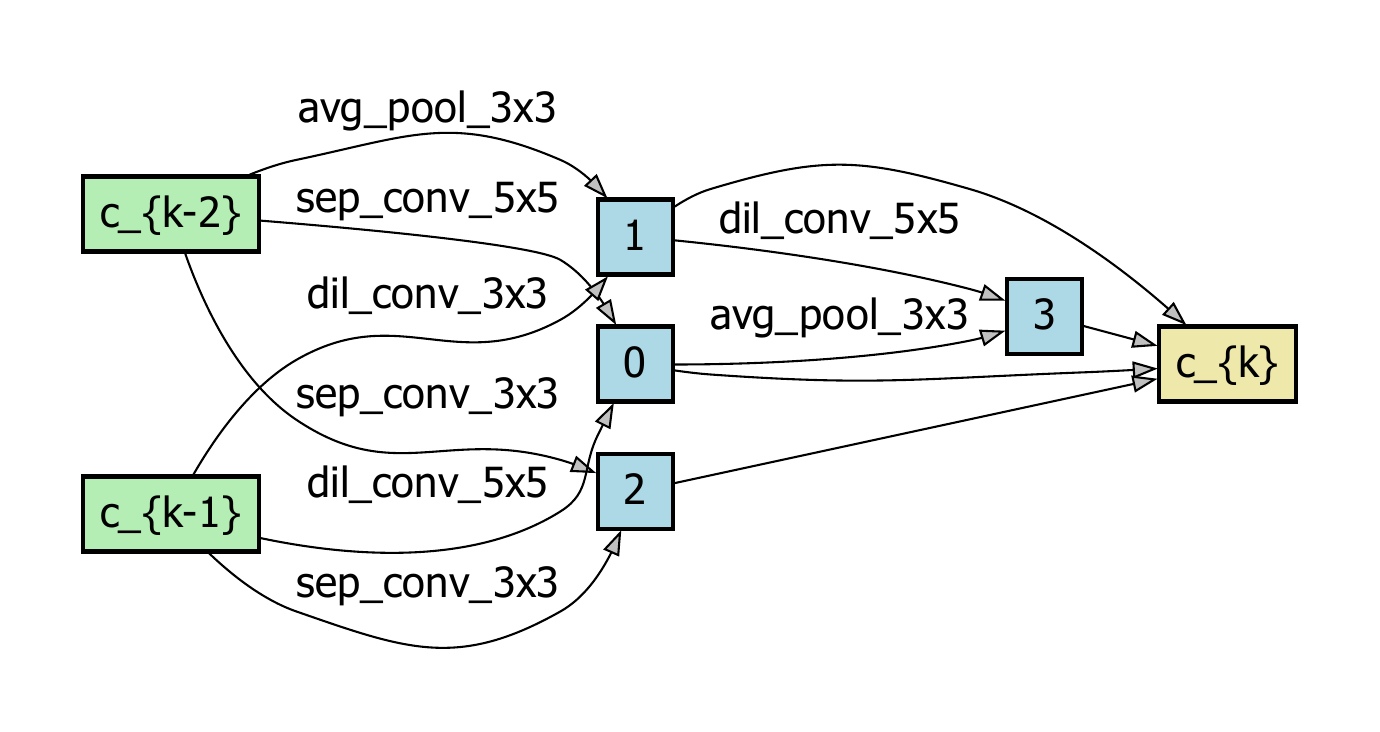}
    \caption{Normal cell on CIFAR-100.}
    \label{fig: cifar100_noraml}
\end{figure}

\begin{figure}[!ht]
    \centering
    \includegraphics[width=0.8\linewidth]{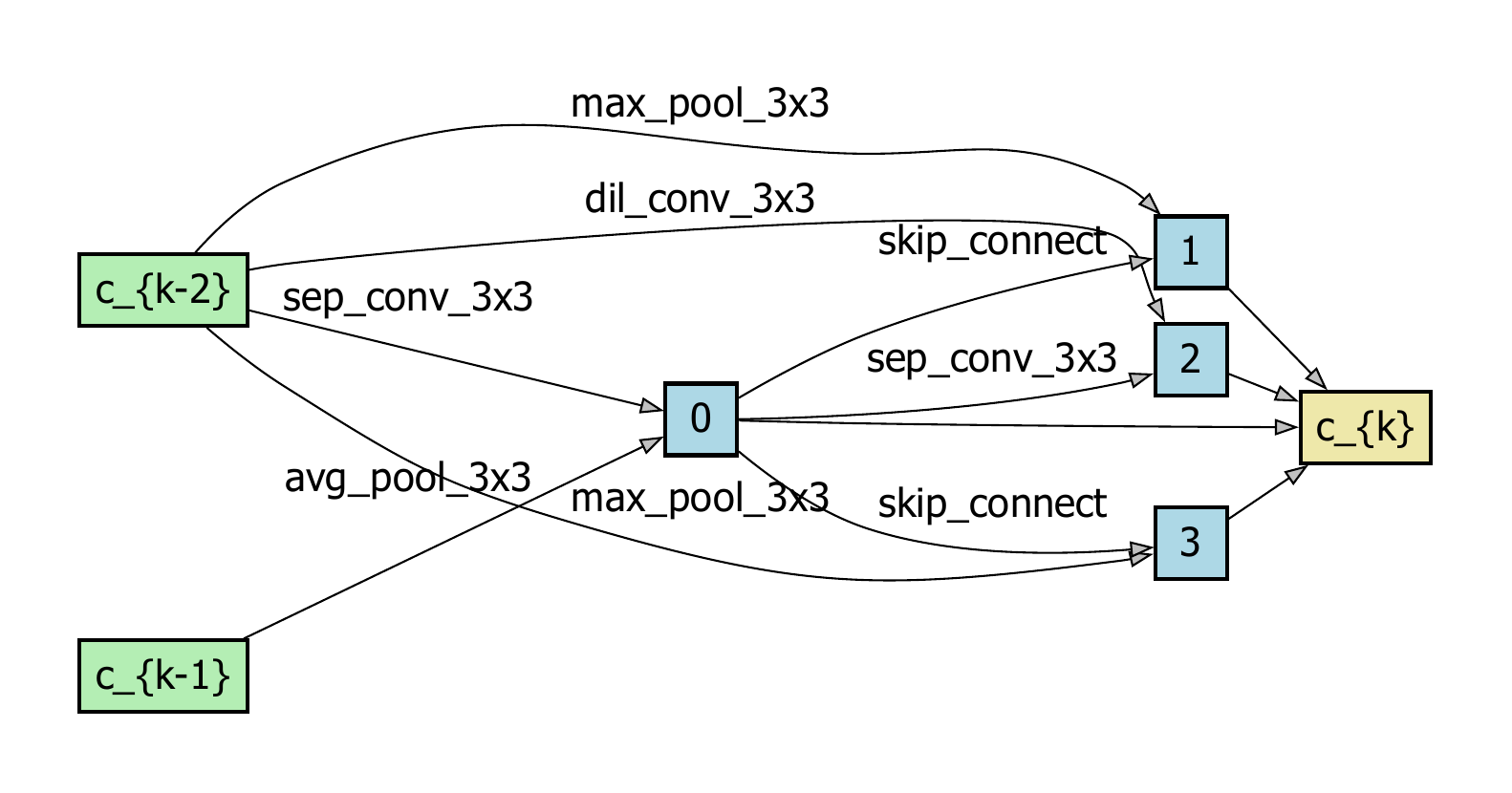}
    \caption{Reduction cell on CIFAR-100.}
    \label{fig: cifar100_reduction}
\end{figure}

We visualize the searched cells on CIFAR-10 and CIFAR-100 (the cells found on ImageNet is the same as on CIFAR-10) in DARTS search space. Fig. \ref{fig: cifar10_noraml} and Fig. \ref{fig: cifar10_reduction} are the searched normal cell and reduction cell on CIFAR-10, and Fig. \ref{fig: cifar100_noraml} and Fig. \ref{fig: cifar100_reduction} are the searched normal cell and reduction cell on CIFAR-100 respectively.

\end{document}